\documentclass[11pt]{article}

\usepackage[preprint]{acl}

\usepackage{times}
\usepackage{latexsym}
\usepackage{pifont}
\usepackage[T1]{fontenc}
\usepackage[utf8]{inputenc}

\usepackage{microtype}

\usepackage{fvextra}
\usepackage{inconsolata}

\usepackage{graphicx}
\usepackage{booktabs}
\usepackage{amsmath}

\usepackage{CJKutf8}

\title{EverMemOS: A Self-Organizing Memory Operating System for Structured Long-Horizon Reasoning}

\author{
  \textbf{Chuanrui Hu}\textsuperscript{1,2}\thanks{~~Equal contribution.} , 
  \textbf{Xingze Gao}\textsuperscript{1,2}\footnotemark[1] , 
  \textbf{Zuyi Zhou}\textsuperscript{1,2}, 
  \textbf{Dannong Xu}\textsuperscript{1,2}, 
  \textbf{Yi Bai}\textsuperscript{1,2}, 
  \textbf{Xintong Li}\textsuperscript{1,2}, 
  \\
  \textbf{Hui Zhang\textsuperscript{1,2}}, 
  \textbf{Tong Li\textsuperscript{1,2}}, 
  \textbf{Chong Zhang\textsuperscript{2}}, 
  \textbf{Lidong Bing\textsuperscript{2}}\thanks{~~Corresponding author.} , 
  \textbf{Yafeng Deng\textsuperscript{1,2}}\footnotemark[2]
  \\
  \textsuperscript{1}EverMind \quad \textsuperscript{2}Shanda Group \\
    \texttt{\{chuanrui.hu, xingze.gao, zuyi.zhou, dannong.xu, baiyi, xintong.li,} \\
    \texttt{zhanghui, litong02, zhangchong, lidong.bing, dengyafeng\}@shanda.com}
}

\usepackage{multirow}

\begin{document}
  \maketitle
\begin{abstract}
Large Language Models (LLMs) are increasingly deployed as long-term interactive agents, yet their limited context windows make it difficult to sustain coherent behavior over extended interactions. Existing memory systems for LLMs often store isolated records and retrieve fragments, limiting their ability to consolidate evolving experience and resolve conflicts. We introduce \textbf{EverMemOS}, a self-organizing memory operating system that implements an engram-inspired lifecycle for computational memory. First, \textit{Episodic Trace Formation} converts dialogue streams into \textit{MemCells} that capture episodic traces, atomic facts, and time-bounded foresight. Second, \textit{Semantic Consolidation} organizes MemCells into thematic \textit{MemScenes}, distilling stable semantic structures and updating user profiles. Finally, \textit{Reconstructive Recollection} performs \textit{MemScene}-guided agentic retrieval to compose the necessary and sufficient context for downstream reasoning. Experiments on LoCoMo, LongMemEval, and PersonaMem-v2 show that EverMemOS significantly outperforms state-of-the-art methods on memory-augmented reasoning tasks. 
%Further profile study on  and qualitative case studies illustrate the chat-oriented capabilities of EverMemOS, including user profiling and foresight. 
Our code is available at \url{https://github.com/EverMind-AI/EverMemOS}.
\end{abstract}

    \begin{figure}[t]
        \centering
        \includegraphics[width=0.48\textwidth]{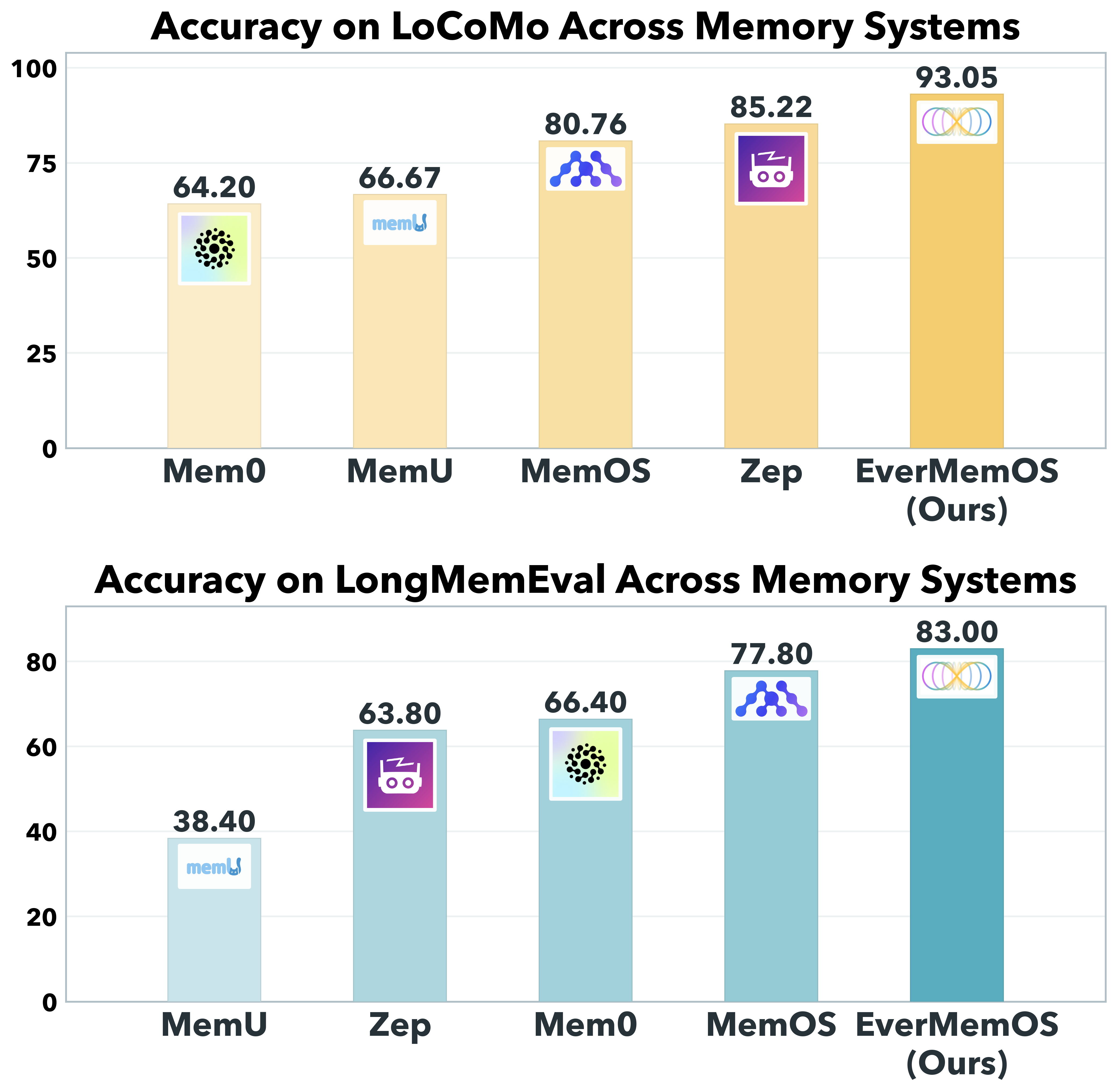}
        % \caption{Comparison of typical fragment-based memory and EverMemOS in an interactive chat scenario.}
        \vspace{-2mm}
        \caption{Evaluation results of different memory methods for LLMs on two benchmarks (LoCoMo and LongMemEval). All methods are based on GPT-4.1-mini.}
        \label{fig:fig1}
        \vspace{-4mm}
    \end{figure}\

    \vspace{-4mm}
  \section{Introduction}

Large Language Models (LLMs) are increasingly deployed as long-term interactive agents rather than transient conversational tools \cite{Yehudai2025LLMAgentEval, Ferrag2025LLMAgents}. For providing better personalized services, LLM-based agents must maintain consistent personas and user models over extended interactions while continuously incorporating new constraints over extended timeframes, spanning days, months, or even years. To address this challenge, expanding context windows is a direct approach, but ultra-long contexts still degrade in performance (e.g., the ``Lost-in-the-Middle'' phenomenon) and incur prohibitive computational costs \cite{liu2024lost}. Consequently, recent research has increasingly focused on constructing memory for LLMs that can both store past information and organize experiences into coherent, evolving structures that support long-horizon reasoning \cite{wu2025long,maharana2024locomo}.

 Recently, a broad range of memory-augmented approaches have been proposed, including retrieval-based memory \cite{zhong2024memorybank,packer2024memgpt}, trainable memory \cite{zheng2024memoryllm,gong2024mplus}, and more recently \textit{Memory Operating Systems} that unify storage, retrieval, filtering, and updating \cite{li2025memos,kang2025memory}. However, enabling long-term consistency in reasoning remains challenging. While these methods improve scalability and modularity, most of them treat memory as flat collections of isolated records. 
As a result, many failures stem not from missing information but from poor integration, where fragmented experiences are not consolidated into higher-level semantic structures. Without consolidation and abstraction, agents may retrieve relevant facts yet fail to detect conflicts, maintain stable user models, or reason consistently over time.
\textbf{Therefore, a key limitation of existing memory methods is the absence of an explicit mechanism to transform fragmented episodic experiences into coherent and stable knowledge structures that support long-horizon reasoning.}

% \bing{If change Figure 1 to a bar chart, need to merge this and the next paragraphs.}
% Motivated by the above observation, we argue that effective long-term interaction requires memory to be designed around an explicit lifecycle, rather than isolated memory operations. Inspired by organizational principles observed in biological memory systems, we propose \textbf{EverMemOS}, a unified Memory Operating System that models memory as a dynamic lifecycle spanning episodic trace formation, semantic consolidation, and reconstructive recollection. This lifecycle-centric view moves beyond recalling individual facts to integrating experiences into structured representations that support consistent reasoning over time.
% %

To address the above limitation, we propose \textbf{EverMemOS}, a unified and product-ready Memory Operating System that models memory as a dynamic lifecycle for long-term LLM-based agents. 
As shown in Figure \ref{fig:fig1}, EverMemOS significantly outperforms the state-of-the-art memory methods for LLMs in experimental evaluation, relatively improving overall accuracy by 9.2\% on LoCoMo and 6.7\% on LongMemEval compared to the strongest baseline method.
EverMemOS aims to transform fragmented episodic experiences into coherent and stable knowledge structures that support long-horizon reasoning through three phases. First, \textbf{Episodic Trace Formation} transforms the unbounded stream of interaction history into discrete, stable memory traces (termed \textit{MemCells}). Second, \textbf{Semantic Consolidation} transforms MemCells into stable, scene-level structures (termed \textit{MemScenes}) that support coherent aggregation, such as maintaining consistent user profiles across interactions. Finally, \textbf{Reconstructive Recollection}, guided by the principle of \textit{necessity and sufficiency}, actively composes only the grounded context required for a given query and supports long-horizon reasoning, rather than indiscriminately retrieving all potentially relevant records.

\begin{figure}[t]
        \centering
        \includegraphics[width=0.45\textwidth]{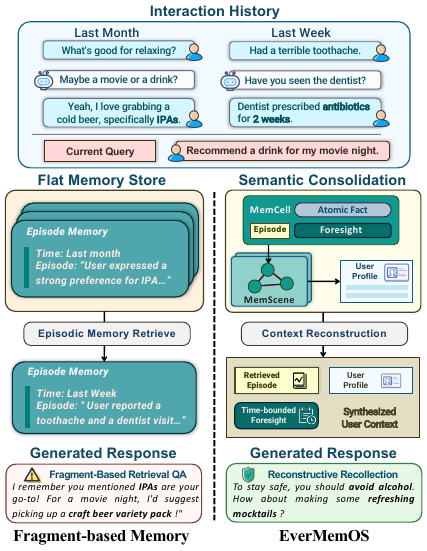}
        \vspace{-2mm}
        \caption{Comparison of typical fragment-based memory and EverMemOS in an interactive chat scenario.}
        \label{fig:comparison}
        \vspace{-4mm}
    \end{figure}

EverMemOS does not aim to simulate biological memory at the neural level. Instead, it draws on organizing principles from biological memory systems and translates them into a computational framework. %In particular, we operationalize the consolidation of transient episodic experiences into stable semantic structures \cite{mcclelland1995there, frankland2005organization}. This lifecycle-based design provides an extensible foundation for robust long-horizon interaction while remaining agnostic to specific downstream tasks or domains. 
%In this work, we focus our quantitative evaluation on memory-augmented reasoning and illustrate profile consistency and foresight-related behaviors through qualitative analysis. 
Figure~\ref{fig:comparison} illustrates the intuition behind EverMemOS. A fragment-based system may recall a user's preference for IPA and recommend an alcoholic drink, failing to account for a newly introduced constraint that the user is taking antibiotics. In contrast, EverMemOS consolidates these experiences into a coherent representation of the user's state, enabling the agent to safely recommend a non-alcoholic alternative. Although such foresight-oriented behaviors are not explicitly captured by existing benchmarks, they expose a fundamental limitation of fragment-based memory and motivate the system-level design of EverMemOS.
Empirically, comprehensive experiments on three benchmarks for memory-augmented reasoning consistently indicate the superiority of EverMemOS, compared to the state-of-the-art methods.
%EverMemOS sets a new state of the art in experimental evaluation, relatively improving overall accuracy by 9.2\% on LoCoMo and 6.7\% on LongMemEval compared to the strongest baseline method. 

Our contributions are summarized as follows:
\begin{itemize}
    \item \textbf{System Design:} We introduce \textbf{EverMemOS}, a unified and product-ready Memory Operating System for LLMs that reconceptualizes memory as a lifecycle, shifting from passive storage of records to structured organization of experience.
    \item \textbf{Innovative Method:} We propose a three-phase method 
    %composed of episodic trace formation, semantic consolidation, and reconstruction recollection. Our method 
    %
    that can transform fragmented episodic experiences into coherent and stable knowledge structures that support long-horizon reasoning.
    \item \textbf{Empirical Validation:} Experimental results demonstrate that EverMemOS achieves state-of-the-art performance on multiple long-context benchmarks for memory-augmented reasoning, validating the effectiveness of lifecycle-based memory organization.
\end{itemize}

  \section{Related Work}

\subsection{Memory Mechanisms in LLMs}

\textbf{Context Window Extension.} Large language models (LLMs) are constrained by fixed-length context windows. Prior work extends context via sparse attention \cite{beltagy2020longformer,zaheer2020bigbird}, recurrence \cite{dai2019transformer,bulatov2022recurrent}, and length extrapolation \cite{chen2024clex,chen2025longpo}. However, longer context does not guarantee effective utilization: the ``Lost-in-the-Middle'' phenomenon persists \cite{liu2024lost, Bulatov2023ScalingTT}, suggesting context extension alone is insufficient for durable memory.

\textbf{Retrieval-Augmented and Parametric Memory.} Retrieval-augmented generation (RAG) \cite{lewis2020retrieval} externalizes memory to alleviate window limits, but its reliability depends on retrieval quality \cite{ram2023incontext}. Parametric approaches internalize information, yet often suffer from forgetting and instability \cite{delange2022continual}. Hybrid approaches \cite{wang2023longmem,packer2024memgpt} alleviate issues but lack a unified organizational principle for persistent memory.

\subsection{Memory Systems}
\textbf{Early Computational Memory.} Early differentiable memory systems (e.g., NTM/DNC/Key--Value memories) \cite{graves2014ntm,graves2016dnc,miller2016kv} introduced external memory interaction, but scale poorly and are ill-suited to modern autoregressive LLMs.

\textbf{Memory in LLM Agents.} As LLM-based agents evolve \cite{xi2023rise, xia2024agent}, memory systems have shifted toward persistent state integration. Recent systems introduce episodic \cite{wang2025mirix}, semantic \cite{shinn2024reflexion}, and hierarchical task memory \cite{sun2025hierarchical}. However, many designs still rely on fragmented text units and limited consolidation, which can degrade long-horizon performance \cite{packer2024memgpt}.

\textbf{Memory Operating Systems.} Recent work formalizes memory management as a system-level runtime. Some focus on lifecycle and capacity, such as \textit{Nemori}'s \cite{nemori2025} prediction-driven updates and \textit{MemoryOS}'s \cite{kang2025memory} hierarchical control. Others, like \textit{Mem0} \cite{mem02025} and \textit{Zep} \cite{zep2025}, prioritize structured fact maintenance via knowledge graphs, while \textit{MemOS} \cite{li2025memos} targets unified scheduling across memory types.

While these systems advance structural organization, they primarily focus on \textit{storage optimization} or \textit{fact maintenance}. EverMemOS distinguishes itself by implementing a \textit{three-phase memory lifecycle} that transforms episodic traces into synthesized semantic structures for long-horizon reasoning.

  \section{EverMemOS}

\subsection{Framework Overview}

\begin{figure*}[t]
    \centering
    \includegraphics[width=0.93\textwidth]{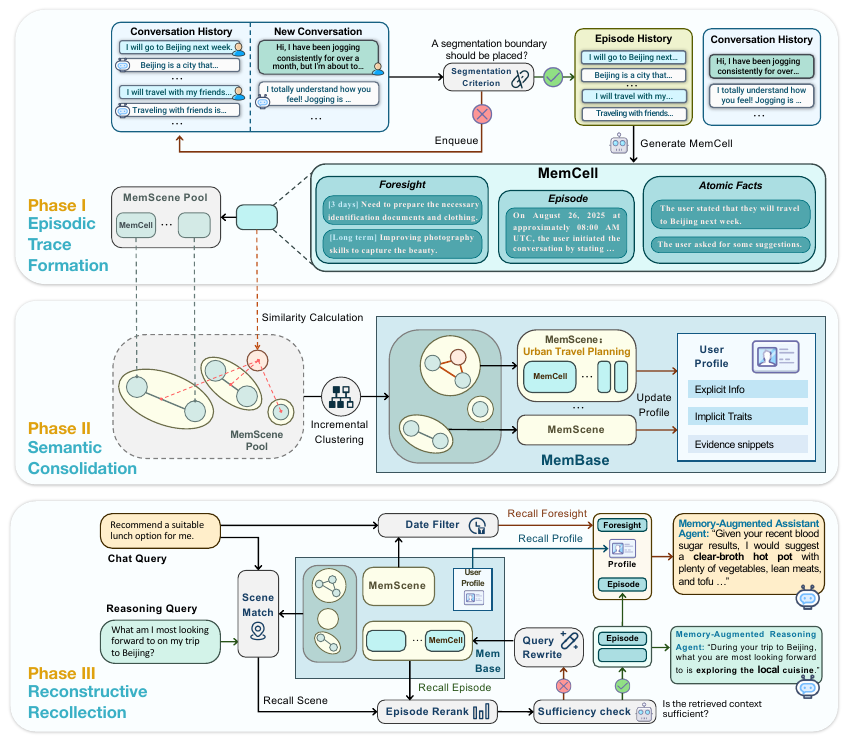}
    \vspace{-2mm}
    \caption{The EverMemOS workflow mirrors an engram-inspired memory lifecycle: (1) \textbf{Episodic Trace Formation} segments continuous dialogue into \textit{MemCells} with episodes, atomic facts, and time-bounded foresight. (2) \textbf{Semantic Consolidation} organizes MemCells into \textit{MemScenes} and updates a user profile. (3) \textbf{Reconstructive Recollection} performs \textit{MemScene}-guided retrieval to compose the necessary and sufficient context.}
    \vspace{-4mm}
    \label{fig:workflow}
\end{figure*}

Drawing inspiration from the biological engram lifecycle \cite{josselyn2015finding}, EverMemOS follows a three-phase workflow (Figure~\ref{fig:workflow}): (1) \textbf{Episodic Trace Formation} encodes interaction streams into \textit{MemCells}; (2) \textbf{Semantic Consolidation} organizes MemCells into \textit{MemScenes} and updates user profiles; and (3) \textbf{Reconstructive Recollection} performs \textit{MemScene}-guided retrieval under the principle of necessity and sufficiency.

\subsection{Memory Primitives}
\label{sec:primitives}

At the core of EverMemOS is the \textbf{MemCell}, the atomic unit bridging low-level data and high-level semantics. Formally, a MemCell $c$ is a tuple $c = (E, \mathcal{F}, P, M)$, where:
\begin{itemize}
    \item $E$ (\textbf{Episode}): A concise third-person narrative of the event, serving as the semantic anchor.
    \item $\mathcal{F} = \{f_1, \dots, f_n\}$ (\textbf{Atomic Facts}): Discrete, verifiable statements derived from $E$ for high-precision matching.
    \item $P$ (\textbf{Foresight}): Forward-looking inferences (prospections; e.g., plans and temporary states) annotated with validity intervals $[t_{start}, t_{end}]$ to support temporal awareness.
    \item $M$ (\textbf{Metadata}): Contextual grounding including timestamps and source pointers.
\end{itemize}
This structure turns memory from a static record ($E, \mathcal{F}$) into a temporally grounded representation that also supports \textbf{Foresight} ($P$).

\subsection{Phase I: Episodic Trace Formation}
Grounded in the engram concept \cite{josselyn2015finding}, this first phase transforms the unbounded stream of interaction history $\mathcal{D}=\{d_1,\ldots,d_T\}$ into discrete, stable memory traces (MemCells). This process adopts a three-step pipeline to distill semantic signal from noisy interaction data:

\paragraph{Contextual Segmentation} To discretize continuous streams, a \textbf{Semantic Boundary Detector} processes interactions via a sliding window. Upon detecting a topic shift, accumulated turns are encapsulated as a raw \textit{episode history}. We implement this step via LLM prompting; while boundary detection is not perfect, we find it robust in downstream evaluation (see Table~\ref{tab:locomo-bound}).

\paragraph{Narrative Synthesis} To resolve dialogue redundancy and ambiguity, the episode history is synthesized into a high-fidelity \textbf{Episode} ($E$). This rewriting process produces a concise, third-person narrative with resolved coreferences, establishing a stable semantic anchor.

\paragraph{Structural Derivation} From $E$, the system extracts \textbf{Atomic Facts} ($\mathcal{F}$) for precise matching and generates \textbf{Foresight} signals ($P$) with inferred validity intervals (e.g., distinguishing temporary "flu" from permanent "graduation"). Concretely, we prompt the LLM over the rewritten Episode $E$ to output a constrained schema of Atomic Facts and Foresight signals with validity intervals $[t_{start}, t_{end}]$. These components are bundled with metadata $M$ to form the final MemCell $c$.

\subsection{Phase II: Semantic Consolidation}
Inspired by systems consolidation \cite{mcgaugh2000memory}, EverMemOS employs an online mechanism that organizes MemCells into higher-order structures to transition from transient episodes to stable long-term knowledge.

\paragraph{Incremental Semantic Clustering}
EverMemOS organizes memory dynamically. When a new MemCell $c$ arrives, the system computes its embedding and retrieves the nearest \textbf{MemScene} centroid. If similarity exceeds a threshold $\tau$, $c$ is assimilated and the scene representation is incrementally updated; otherwise, a new \textbf{MemScene} is instantiated. This online process maintains thematic structure in real-time without batch reprocessing.

\paragraph{Scene-Driven Profile Evolution}
Scene-level consolidation can also update a compact \textbf{User Profile} from aggregated evidence. When a new MemCell is assimilated into a \textbf{MemScene}, EverMemOS updates a concise scene summary and refreshes the user profile by prompting over these summaries (rather than individual turns), helping separate stable traits from temporary states. We maintain a compact profile of \emph{explicit facts} (including time-varying measurements) and \emph{implicit traits}, updated online from scene summaries with recency-aware updates and conflict tracking (Appendix~\ref{sec:profile_example}).

\subsection{Phase III: Reconstructive Recollection}
Building on theories of reconstructive memory \cite{schacter2008searching}, retrieval in EverMemOS is modeled not as a static lookup but as an active \textbf{Reconstruction} process, guided by the principle of \textit{necessity and sufficiency}. Given a query $q$, EverMemOS performs \textbf{agentic retrieval} grounded in \textit{MemScenes}.

\paragraph{MemScene Selection} We first compute relevance between the query and all MemCells by fusing dense and BM25 retrieval over their Atomic Facts $\mathcal{F}$ via Reciprocal Rank Fusion (RRF). We then score each MemScene by the maximum relevance among its constituent MemCells and select a small set of the highest-scoring MemScenes.

\paragraph{Episode and Foresight Filtering} Within the selected MemScenes, we pool Episodes from their constituent MemCells and \textbf{re-rank} them to select a compact set for downstream inference. We then apply \textbf{Foresight Filtering}, retaining only time-valid Foresight whose validity intervals satisfy $t_{now}\in[t_{start}, t_{end}]$ (discarding expired ones).

\paragraph{Agentic Verification and Query Rewriting} The retrieved context is evaluated by an LLM-based verifier for sufficiency. If it is deemed insufficient, the system triggers a \textbf{query rewriting} step to supplement retrieval; otherwise, the context is passed to the downstream module. Prompt templates are provided in Appendix~\ref{sec:agentic_prompts}.

\paragraph{Task Modes} We consider two downstream settings that share the same retrieval pipeline: \textbf{Memory-Augmented Reasoning} and \textbf{Memory-Augmented Chat}. For \textit{Reasoning}, we use the retrieved Episodes as context for benchmark evaluation. For \textit{Chat}, the composed context additionally incorporates the User Profile and time-valid Foresight signals, filtered by the current time $t_{now}\in[t_{start}, t_{end}]$; since these capabilities are not covered by existing reasoning benchmarks, we present them through qualitative case studies.

  \section{Experiments}
%\bing{图表在论文中的位置，原则上尽量贴近主要讨论他们的文本部分，最后排版的时候注意一下}
We evaluate EverMemOS on two long-horizon memory-augmented reasoning benchmarks (LoCoMo~\cite{maharana2024locomo} and LongMemEval~\cite{wu2025long}), and report a profile study on PersonaMem-v2~\cite{jiang2025personamem}.

\subsection{Experimental Setup}

\paragraph{Benchmarks}
We evaluate memory-augmented reasoning on LoCoMo and LongMemEval. LoCoMo contains 1{,}540 questions over 10 ultra-long dialogues ($\sim$9K tokens each), spanning single-hop, multi-hop, and temporal questions. LongMemEval (S-setting, $\sim$115k tokens per conversation) evaluates 500 questions requiring full-history parsing across core capabilities (e.g., updates and abstention). We additionally evaluate user profiling on PersonaMem-v2.

\paragraph{Baselines}
We compare EverMemOS against state-of-the-art memory systems: \textbf{Zep} \cite{zep2025}, \textbf{Mem0} \cite{mem02025}, \textbf{MemOS} \cite{li2025memos}, \textbf{MemoryOS} \cite{kang2025memory}, and \textbf{MemU}\footnote{Open-source memory infrastructure: \url{https://github.com/NevaMind-AI/memU}}.
\textbf{Fair comparison:} We standardize the answer-generation backbone across methods while keeping each baseline's official memory configuration unchanged; for LongMemEval, we report baseline scores from the official \textbf{MemOS} leaderboard. Full settings are provided in Appendix~\ref{sec:eval_setting}.

\paragraph{Evaluation Protocol}
We adopt the \textbf{LLM-as-a-judge} protocol, following MemOS: each answer is evaluated by GPT-4o-mini and two auxiliary judge models, and scores are averaged across the three judgments in a \emph{blind} setting. We validate the reliability of this protocol against human annotations in Section~\ref{sec:judge_reliability} (Appendix), showing high agreement (Cohen's $\kappa > 0.89$).

\paragraph{Implementation Details}
EverMemOS uses GPT-4.1-mini (or GPT-4o-mini where specified) for all reasoning and memory operations. Retrieval uses hybrid dense+BM25 fusion (RRF) with re-ranking. Default retrieval hyperparameters are in Appendix~\ref{sec:eval_setting}. Unless otherwise specified, quantitative experiments use \textbf{Memory-Augmented Reasoning}.
We provide a token-level cost breakdown by lifecycle phase in Appendix (Table~\ref{tab:token_cost}).

\subsection{Main Results}

\begin{table*}[!h]
\centering
\small
\setlength{\tabcolsep}{5pt}
\begin{tabular}{l r ccccc}
\toprule
\textbf{Method} & \textbf{Avg. Tokens} & \textbf{Single Hop} & \textbf{Multi Hop} & \textbf{Temporal} & \textbf{Open Domain} & \textbf{Overall} \\
\midrule
\multicolumn{7}{l}{\textit{GPT-4o-mini backbone}} \\[-2pt]
MemoryOS & 5.2k & 62.43 & 56.50 & 37.18 & 40.28 & 54.70 \\
Mem0 & 1.0k & 66.71 & 58.16 & 55.45 & 40.62 & 61.00 \\
MemU & 4.0k & 72.77 & 62.41 & 33.96 & 46.88 & 61.15 \\
MemOS & 2.5k & 81.45 & 69.15 & 72.27 & 60.42 & 75.87 \\
Zep & 1.4k & 88.11 & 71.99 & 74.45 & 66.67 & 81.06 \\
\textbf{EverMemOS} & 2.5k & \textbf{91.08} {\scriptsize ($\uparrow$3.4\%)} & \textbf{86.17} {\scriptsize ($\uparrow$19.7\%)} & \textbf{81.93} {\scriptsize ($\uparrow$10.0\%)} & \textbf{66.67} {\scriptsize ($\uparrow$0.0\%)} & \textbf{86.76} {\scriptsize ($\uparrow$7.0\%)} \\

\midrule
\multicolumn{7}{l}{\textit{GPT-4.1-mini backbone}} \\[-2pt]
MemoryOS & 5.5k & 67.30 & 59.34 & 42.26 & 59.03 & 60.11 \\
Mem0 & 1.0k & 68.97 & 61.70 & 58.26 & 50.00 & 64.20 \\
MemU & 4.0k & 74.91 & 72.34 & 43.61 & 54.17 & 66.67 \\
MemOS & 2.5k & 85.37 & 79.43 & 75.08 & 64.58 & 80.76 \\
Zep & 1.4k & 90.84 & 81.91 & 77.26 & 75.00 & 85.22 \\
\textbf{EverMemOS} & 2.3k & \textbf{96.67} {\scriptsize ($\uparrow$6.4\%)} & \textbf{91.84} {\scriptsize ($\uparrow$12.1\%)} & \textbf{89.72} {\scriptsize ($\uparrow$16.1\%)} & \textbf{76.04} {\scriptsize ($\uparrow$1.4\%)} & \textbf{93.05} {\scriptsize ($\uparrow$9.2\%)} \\
\bottomrule
\end{tabular}
\caption{Main results on \textbf{LoCoMo} under two backbones. All metrics are accuracy (\%), except Avg. Tokens. For EverMemOS, values in parentheses denote relative change (\%) compared to the strongest baseline under the same backbone.}
\label{tab:locomo-main}
\end{table*}
\begin{table*}[!h]
    \centering
    \small
    \setlength{\tabcolsep}{2pt}
    \begin{tabular}{@{}l c ccccccc@{}}
    \toprule
    \textbf{Method} & \textbf{Token} & \textbf{SS-User} & \textbf{SS-Asst} & \textbf{SS-Pref} & \textbf{Multi-S} & \textbf{Know. Upd} & \textbf{Temp. Reas} & \textbf{Overall} \\
    \midrule
    MemU & 0.5k & 67.14 & 19.64 & 76.67 & 42.10 & 41.02 & 17.29 & 38.40 \\
    Zep & 1.6k & 92.90 & 75.00 & 53.30 & 47.40 & 74.40 & 54.10 & 63.80 \\
    Mem0 & 1.1k & 82.86 & 26.78 & 90.00 & 63.15 & 66.67 & 72.18 & 66.40 \\
    MemOS & 1.4k & 95.71 & 67.86 & \textbf{96.67} & 70.67 & 74.26 & \textbf{77.44} & 77.80 \\
    \textbf{EverMemOS} & 2.8k & \textbf{97.14} {\tiny ($\uparrow$1.5\%)} & \textbf{85.71} {\tiny ($\uparrow$14.3\%)} & 93.33 {\tiny ($\downarrow$3.5\%)} & \textbf{73.68} {\tiny ($\uparrow$4.3\%)} & \textbf{89.74} {\tiny ($\uparrow$20.6\%)} & \textbf{77.44} {\tiny ($\uparrow$0.0\%)} & \textbf{83.00} {\tiny ($\uparrow$6.7\%)} \\
    \bottomrule
    \end{tabular}
    \caption{Main results on \textbf{LongMemEval} (accuracy, \%). SS denotes single-session tasks; baselines are from the official \textbf{MemOS} results \cite{li2025memos}. For EverMemOS, values in parentheses denote relative change (\%) compared to the strongest baseline for that metric.}
    \label{tab:longmemeval-main}
\end{table*}

Main results on two benchmarks are reported in Tables~\ref{tab:locomo-main}-\ref{tab:longmemeval-main}. We make three observations:

(1) \textbf{Lifecycle-driven performance gains.}
EverMemOS outperforms the strongest baseline on each benchmark overall, i.e., Zep on LoCoMo by 7.0\% and 9.2\%, and MemOS on LongMemEval by 6.7\%.
We attribute this to the shift from flat memory storage to a structured lifecycle, which consolidates fragmented experiences into usable knowledge before retrieval, providing a more robust context than isolated record matching.

(2) \textbf{Structural consolidation aids complex reasoning that requires integrating dispersed evidence.}
%Improvements are most pronounced on multi-hop or temporal questions in LoCoMo and on knowledge update or SS-Asst in LongMemEval, while open domain remains comparable to Zep. 
% \bing{对两个benchmarks的评测维度，写法要一直，比如首字母大小写、是否加粗}\
We can observe significant gains on LoCoMo multi-hop (+19.7\%) and temporal (+10.0\%) tasks, as well as LongMemEval knowledge update (+20.6\%), validating the effectiveness of MemScenes. By clustering related episodes into coherent thematic units, EverMemOS presents the solver with a complete narrative context. This enables LLMs to naturally bridge dispersed evidence and resolve state conflicts that confuse other models relying on fragmented retrieval.

(3) \textbf{EverMemOS offers a favorable accuracy-efficiency trade-off.}
%Figure~\ref{fig:frontier} shows that EverMemOS attains high accuracy with moderate retrieval budgets, suggesting that improvements come from composing necessary and sufficient evidence rather than overly expanding context.
%
%\textbf{Precision via Reconstructive Recollection.} EverMemOS offers a favorable accuracy-efficiency trade-off. 
As shown in Figure~\ref{fig:frontier} , EverMemOS attains high accuracy with moderate retrieval budgets. This efficiency confirms the utility of the Reconstructive Recollection phase, where the agentic sufficiency check ensures the context is composed of necessary and sufficient evidence, avoiding the noise accumulation common in fixed-budget retrieval.

\subsection{Ablation Study}
\label{sec:ablation}
We conduct ablations on LoCoMo to isolate the contributions of \textit{MemScenes}, \textit{MemCells}, and episode segmentation.

\begin{figure}[t]
    \centering
    \includegraphics[width=0.95\linewidth]{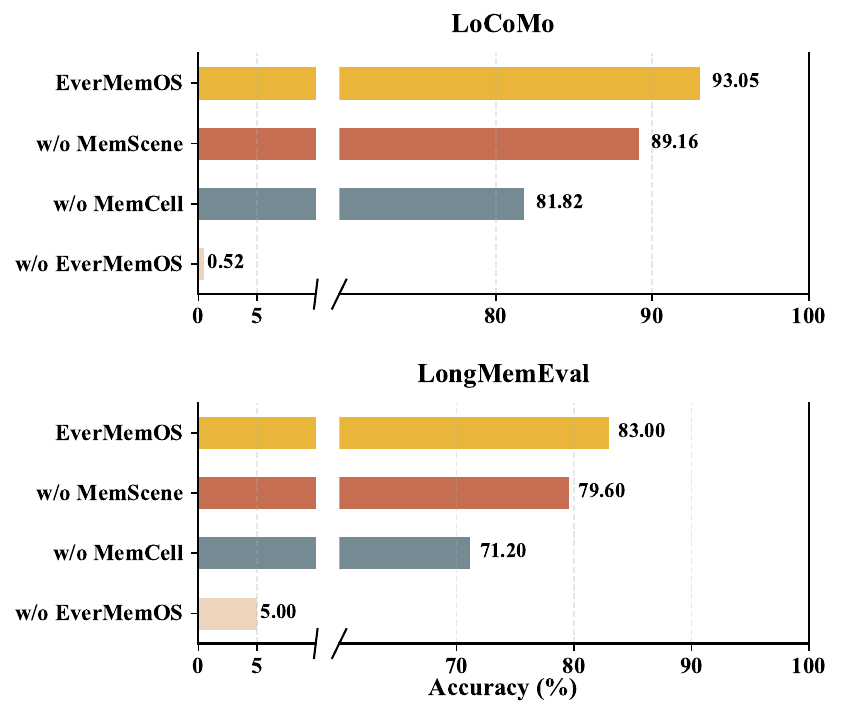}
    \vspace{-2mm}
    \caption{Ablation results (overall accuracy) on LoCoMo and LongMemEval.}
    \vspace{-4mm}
    \label{fig:ablation_chart}
\end{figure}

\begin{figure}[t]
    \centering
    \includegraphics[width=0.9\linewidth]{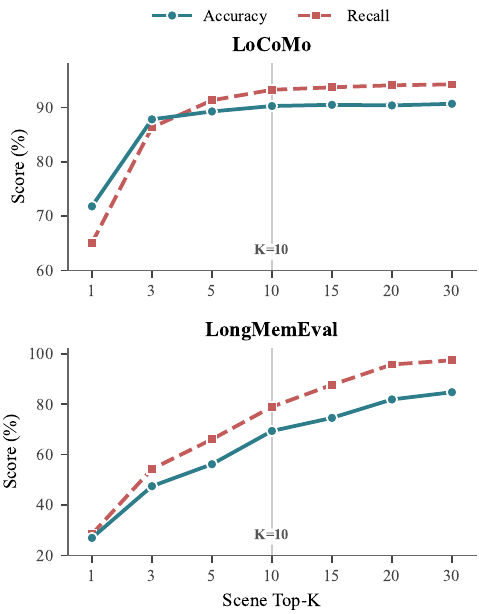}
    \vspace{-2mm}
    \caption{Sensitivity analysis on the MemScene count ($N$).}
    \vspace{-4mm}
    \label{fig:hyperpara}
\end{figure}

\begin{figure}[t]
    \centering
    \includegraphics[width=0.9\linewidth]{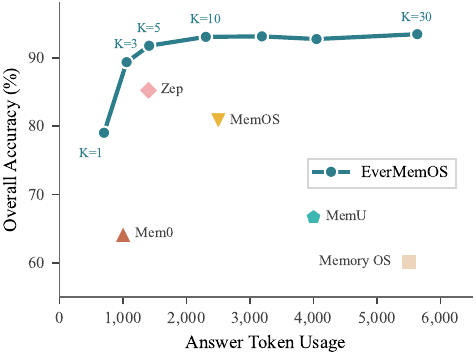}
    \vspace{-2mm}
    \caption{Performance vs. cost frontier on LoCoMo by varying the retrieved episode count ($K$).}
    \vspace{-4mm}
    \label{fig:frontier}
\end{figure}

\begin{figure*}[t]
    \centering
    \includegraphics[width=0.9\linewidth]{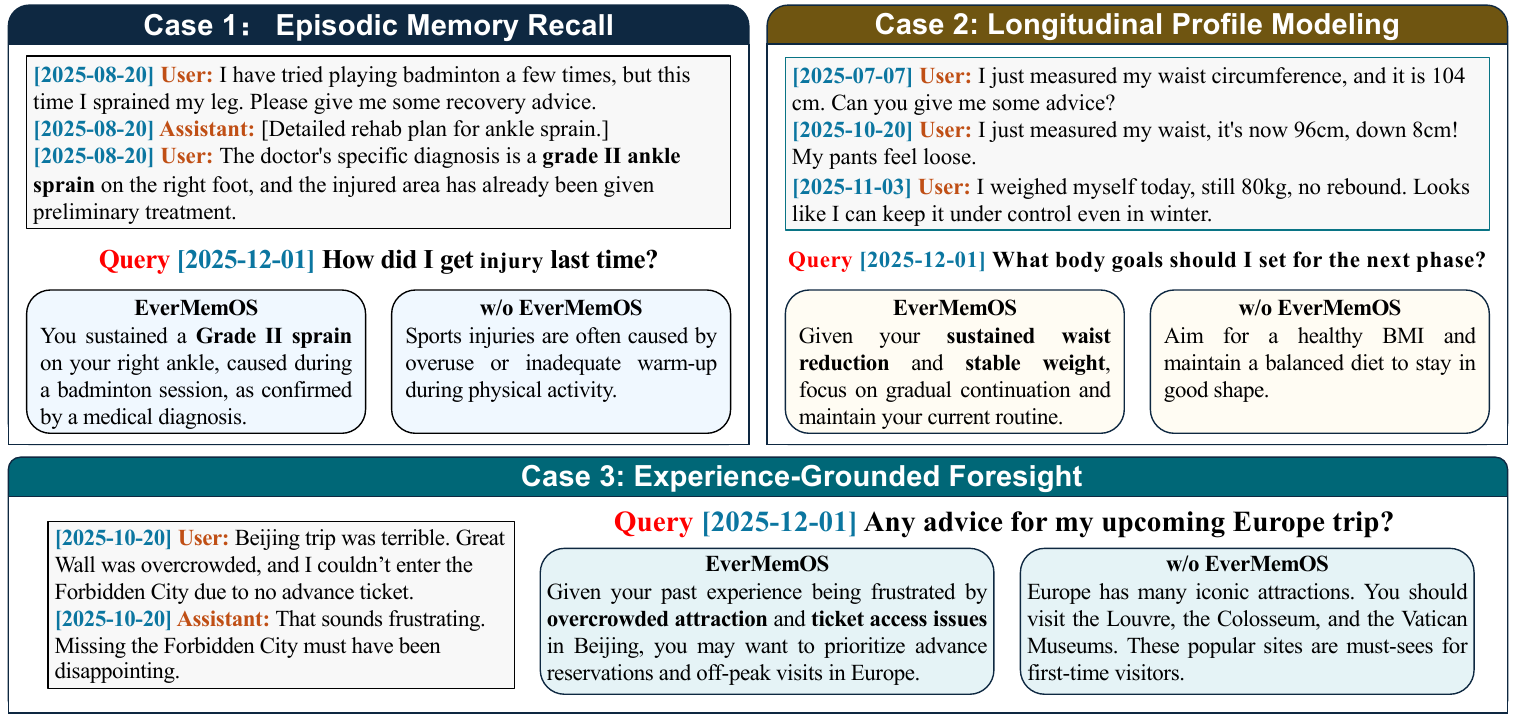}
    \vspace{-2mm}
    \caption{Case studies illustrating Profile, Foresight, and Episode capabilities in Memory-Augmented Chat.}
    \vspace{-2mm}
    \label{fig:case_study}
\end{figure*}

\textbf{Impact of Memory Architecture.} 
To isolate the contribution of memory structure, we compare EverMemOS with three degraded variants: w/o EverMemOS (no external memory), w/o MemScene (flat retrieval over MemCells), and w/o MemCell (retrieval over raw dialogue). The backbone model and prompts are fixed, and only the memory representation and retrieval pipeline are varied.

As shown in Figure~\ref{fig:ablation_chart}, performance degrades stepwise as structure is removed, revealing three corresponding capability losses. Removing \textit{MemScenes} eliminates scene-level organization, weakening cross-turn aggregation over related episodes. Removing \textit{MemCells} further drops the stable semantic units (episodes/facts), forcing retrieval to rely on raw dialogue matching. Finally, removing external memory collapses long-horizon performance, indicating that many queries cannot be handled reliably within the context window alone.

\textbf{Effectiveness of Episode Segmentation.} 
We evaluate semantic episode segmentation against fixed heuristics and ground-truth boundaries under \textbf{w/o MemScene} to isolate boundary quality.
We compare three strategies: (1) \textbf{Fixed Heuristics} (fixed message count $N=10$ or token thresholds $N=512,1024$); (2) \textbf{Session (Oracle)} (ground-truth session boundaries); and (3) \textbf{EverMemOS} (semantic segmentation with different backbones).

\begin{table}[!t]
    \centering
    \small
    \renewcommand{\arraystretch}{1.15}
    \setlength{\tabcolsep}{5pt}
    
    \begin{tabular}{l c c}
        \toprule
        & \multicolumn{2}{c}{\textbf{Answer Model}} \\
        \cmidrule(lr){2-3}
        \textbf{Segmentation Method} & \textbf{GPT-4.1-mini} & \textbf{Qwen3-4B} \\
        \midrule
        \multicolumn{3}{l}{\textit{Heuristic Baselines}} \\
        Fixed-Message-10 & 88.05 & 80.95 \\
        Fixed-Token-512 & 87.55 & 80.67 \\
        Fixed-Token-1024 & 84.52 & 75.19 \\
        \midrule
        \multicolumn{3}{l}{\textit{Semantic Segmentation}} \\
        Session (Oracle) & 87.66 & 80.63 \\
        \textbf{Default (EverMemOS)} &  &  \\
        $\quad$ w/ GPT-4.1-mini & 89.16 & 83.07 \\
        $\quad$ w/ Qwen3-4B & 89.78 & 82.73 \\
        \bottomrule
    \end{tabular}
    \vspace{-2mm}
    \caption{
        Comparison of boundary detection strategies.
        \textbf{Session (Oracle)} uses the ground-truth session partitions provided by LoCoMo.
    }
    \vspace{-4mm}
    \label{tab:locomo-bound}
\end{table}

Table~\ref{tab:locomo-bound} shows that (i) semantic segmentation consistently outperforms fixed heuristics, especially coarse token chunking; (ii) it also outperforms Session (Oracle), suggesting sessions are not always optimal retrieval units; and (iii) results are robust across boundary-detection backbones (accuracy changes $\leq$0.7 points).

\subsection{Hyperparameter Analysis}
\label{sec:sensitivity}
We investigate the impact of retrieval scope via two hyperparameters: the number of retrieved \textit{MemScenes} ($N$) and episodes ($K$). As shown in Figure~\ref{fig:hyperpara}, performance gains saturate around $N=10$. Figure~\ref{fig:frontier} further illustrates the efficiency--accuracy frontier governed by $K$. We therefore adopt $N=10$ and $K=10$ as the default configuration to balance performance with computational cost. Comprehensive sensitivity analysis is detailed in Appendix~\ref{sec:sensitivity_details}.

\subsection{Profile Study}
\label{sec:profile_study}
\begin{table}[t]
    \centering
    \small
    \setlength{\tabcolsep}{3pt}
    \renewcommand{\arraystretch}{1.0}
    \begin{tabular}{l r r r}
    \toprule
    \textbf{Scenario} & \textbf{Ep.+Prof.} & \textbf{Prof.-only} & \textbf{Ep.-only} \\
    \midrule
Consultation        & \textbf{51.03} & 47.33 & 44.44 \\
Email (Personal)    & \textbf{53.85} & 46.15 & 46.15 \\
Translation         & \textbf{50.00} & 46.15 & 38.08 \\
Email (Professional) & \textbf{53.79} & 41.38 & 45.17 \\
Writing (Creative)    & \textbf{55.10} & 48.57 & 42.04 \\
Writing (Professional) & \textbf{45.56} & 44.79 & 40.15 \\
Knowledge Query     & \textbf{63.68} & 62.94 & 54.73 \\
Social Media        & \textbf{47.90} & 44.96 & 36.13 \\
Chat                & \textbf{52.09} & 44.87 & 41.83 \\
\midrule
\textbf{Overall}    & \textbf{53.25} & 48.30 & 43.93 \\
    \bottomrule
    \end{tabular}
    %\vspace{-2mm}
    \caption{Profile ablation on PersonaMem v2~\cite{jiang2025personamem} (5{,}000 questions across 9 scenarios; accuracy, \%).}
    \vspace{-2mm}
    \label{tab:profile_effect}
    \end{table}
We evaluate the effect of the consolidated user profile on PersonaMem-v2 (32k)~\cite{jiang2025personamem}; results are not directly comparable across dataset versions due to differences in task setup and annotations. Table~\ref{tab:profile_effect} shows that adding the User Profile to episodic evidence improves overall accuracy by 9.32 points over episodes-only (53.25 vs.\ 43.93), indicating that semantic consolidation provides complementary signal beyond episodic retrieval. We defer the full comparison against other memory systems on PersonaMem-v2 to Appendix~\ref{sec:personamem_full}.

\subsection{Case Study}
Existing benchmarks primarily evaluate answer-level accuracy/recall and do not capture several capabilities required for long-term conversational agents, such as conflict detection, profile stability, and experience-grounded foresight. To complement quantitative results, Figure~\ref{fig:case_study} shows three representative cases: \textbf{(Episode)} reconstructing a concrete past injury episode (a Grade-II ankle sprain during badminton) rather than producing a generic explanation; \textbf{(Profile)} maintaining longitudinal stability and using sustained improvements (waist 104$\rightarrow$96\,cm with stable weight) for trajectory-consistent goal setting; and \textbf{(Foresight)} leveraging previously observed failures (overcrowding and missing advance tickets) to make proactive recommendations for future travel.
Together, these cases illustrate coherent, experience-aware behavior beyond what is measured by existing benchmarks.

  \section{Conclusion}

In this paper, we introduced EverMemOS, a unified memory operating system for long-horizon LLM agents. By modeling an explicit memory lifecycle composed of episodic trace formation, semantic consolidation, and reconstructive recollection, EverMemOS achieves state-of-the-art performance on memory-augmented reasoning benchmarks, with particularly strong gains on multi-hop and temporal questions. We hope EverMemOS provides an extensible foundation for building more consistent and context-aware interactive agents.

  \section*{Limitations}
We evaluate EverMemOS on text-only conversational benchmarks. Although the \textit{MemCell} and \textit{MemScene} abstraction is modality-agnostic, extending EverMemOS to multimodal or embodied settings is beyond the scope of this work.
EverMemOS introduces LLM-mediated operations for memory construction and retrieval, increasing latency and computational cost relative to single-pass baselines. While many components can be cached, batched, or run asynchronously, improving end-to-end efficiency remains future work.
Finally, current benchmarks lack protocols for stress-testing ultra-long timelines, so our evaluation does not fully isolate performance in such regimes. This motivates future benchmarks for long-term memory organization and consolidation.

  % \bibliographystyle{acl_natbib}
  \bibliography{custom}
  \clearpage
  \appendix

  \section{Evaluation Details}
\label{sec:appendix_eval}

\subsection{Evaluation Settings and Fair Comparison}
\label{sec:eval_setting}

\paragraph{LoCoMo backbones.} We report LoCoMo results under two backbones: GPT-4.1-mini (primary, reflecting an up-to-date backbone) and GPT-4o-mini (to facilitate comparison with prior work). Following common practice in LTMOS evaluation, we standardize the backbone used for \emph{final answer generation} to isolate the contribution of memory management from the base model.

\paragraph{Baseline executability.} For EverMemOS and MemoryOS, we execute the full pipeline (memory construction, retrieval, and answering) with the specified backbone. For Mem0, MemU, MemOS, and Zep, we use their official APIs for memory management/retrieval; in this setting, we keep each baseline's official memory configuration and prompting unchanged and apply the unified backbone only at the answering stage.

\paragraph{LongMemEval.} Due to the extreme input length of LongMemEval, we cannot stably run all baseline APIs end-to-end; we therefore report baseline results from the official MemOS leaderboard\footnote{Leaderboard data: \url{https://huggingface.co/datasets/MemTensor/MemOS_eval_result}} and evaluate EverMemOS with GPT-4.1-mini under the same protocol.

\paragraph{Retrieval configuration.} EverMemOS uses a hybrid retriever that fuses dense retrieval (encoder: \texttt{Qwen3-Embedding-4B}~\cite{qwen3embedding}) and sparse retrieval (BM25) via Reciprocal Rank Fusion (RRF), followed by episode re-ranking (\texttt{Qwen3-Reranker-4B}~\cite{qwen3embedding}). Unless otherwise specified, we retrieve the top-10 \textit{MemScenes} and select 10 Episodes for downstream inference.

\paragraph{MemBase statistics and construction hyperparameters.} We report dataset-level MemBase statistics (Table~\ref{tab:membase_stats}) and memory-construction hyperparameters (Table~\ref{tab:membase_hparams}) for LoCoMo and LongMemEval. \textit{MemScenes} are the clustering units produced by Phase~II, and each MemScene contains a small set of \textit{MemCells}. We use the same pipeline across datasets, while adopting dataset-specific clustering hyperparameters to reflect different dialogue structures and time spans. LongMemEval contains 500 dialogue--question pairs (one per conversation). \textbf{Max time gap} is the maximum allowed temporal distance (in days): when assigning a MemCell $A$ to a candidate MemScene, if the closest-in-time MemCell $B$ already in that MemScene is farther than this threshold, $A$ is not clustered into that MemScene.

\begin{table}[t]
\centering
\footnotesize
\setlength{\tabcolsep}{2pt}
\renewcommand{\arraystretch}{1.0}
\caption{MemBase statistics on LoCoMo and LongMemEval.}
\label{tab:membase_stats}
\begin{tabular}{@{}p{0.55\linewidth} r r@{}}
\toprule
\textbf{Metric} & \textbf{LoCoMo} & \textbf{LongMemEval} \\
\midrule
\multicolumn{3}{l}{\textit{Dataset scale}} \\[-2pt]
\#Conversations & 10 & 500 \\
\#Questions & 1{,}540 & 500 \\
\midrule
\multicolumn{3}{l}{\textit{MemBase statistics}} \\[-2pt]
\#Total MemCells & 702 & 54{,}755 \\
\#Total MemScenes & 286 & 40{,}138 \\
Avg MemCells/conv. & 70.2 & 109.5 \\
Avg MemScenes/conv. & 28.6 & 80.3 \\
Avg MemCells/MemScene & 2.45 & 1.36 \\
MemCells/conv. (range) & 34--95 & 82--154 \\
MemScenes/conv. (range) & 13--49 & 60--102 \\
\bottomrule
\end{tabular}
\end{table}

\begin{table}[t]
\centering
\footnotesize
\setlength{\tabcolsep}{6pt}
\renewcommand{\arraystretch}{1.0}
\caption{Memory-construction hyperparameters.}
\label{tab:membase_hparams}
\begin{tabular}{@{}l c c@{}}
\toprule
\textbf{Hyperparameter} & \textbf{LoCoMo} & \textbf{LongMemEval} \\
\midrule
Clustering threshold $\tau$ & 0.70 & 0.50 \\
Max time gap (days) & 7 & 30 \\
\bottomrule
\end{tabular}
\end{table}

\paragraph{Multi-round query rewriting frequency.} On LoCoMo (GPT-4.1-mini), the sufficiency checker triggers a second-round query rewriting for 31.0\% of questions.

\paragraph{Default evaluation mode.} Unless otherwise specified, quantitative experiments use \textbf{Memory-Augmented Reasoning} (Episodes-only). We additionally report the effect of the consolidated \textbf{Profile} in Table~\ref{tab:profile_effect}, while \textbf{Foresight} is illustrated in the qualitative \textbf{Case Study} (\textbf{Memory-Augmented Chat}).

\subsection{LLM-as-Judge Reliability}
\label{sec:judge_reliability}
We randomly selected 25 non-overlapping Q\&A pairs from LoCoMo and 25 from LongMemEval, and generated model answers for each question. We recruited annotators via Prolific. For each Q\&A pair, five independent human evaluators judged whether the generated answer was correct given the question and the reference answer. All participants provided informed consent via the platform interface and were compensated at approximately \$12.00/hour, consistent with fair-pay guidelines for academic research and above local minimum wage standards. Table~\ref{tab:llm_eval} shows strong agreement between the LLM-as-judge protocol and human annotations: Cohen’s $\kappa$ exceeds 0.89 and accuracy remains above 98\% across benchmarks. Pearson $r$ is 0.891 on LoCoMo and 0.979 on LongMemEval. These results suggest that GPT-4o-mini achieves human-level reliability for answer verification, enabling evaluation that is rigorous, reproducible, and cost-efficient.

\begin{table}[h]
\centering
\caption{Reliability matrix for LLM-as-Judge.}
\label{tab:llm_eval}
\resizebox{\linewidth}{!}{%
  \setlength{\tabcolsep}{6pt}%
  \renewcommand{\arraystretch}{0.9}%
  \begin{tabular}{l c c c c}
    \toprule
    \textbf{Model} & \textbf{Cohen's $\kappa$} & \textbf{95\% CI} & \textbf{Accuracy} & \textbf{Pearson $r$} \\
    \midrule
    LoCoMo               & 0.891 & {[0.742, 1.000]} & 0.984 & 0.891 \\
    LongMemEval          & 0.978 & {[0.936, 1.000]} & 0.992 & 0.979 \\
    \bottomrule
  \end{tabular}%
}
\end{table}

\subsection{Token Cost Breakdown}

To improve cost transparency, we log \emph{all} LLM API calls during LoCoMo evaluation (1,540 questions) under two backbones (GPT-4.1-mini and GPT-4o-mini) and attribute token usage to stages in our pipeline. Since LoCoMo evaluation uses \textbf{Memory-Augmented Reasoning} (Episodes-only), we do \emph{not} invoke the \textbf{Profile} module; therefore, profile-related tokens are excluded from Table~\ref{tab:token_cost}. Table~\ref{tab:token_cost} maps stages to EverMemOS phases. Phase~I corresponds to memory construction (\texttt{add}). In this Episodes-only setting, Phase~II uses non-LLM computation (clustering/embedding updates) and thus incurs no additional LLM tokens. Phase~III consists of retrieval (\texttt{search}) and answer generation (\texttt{answer}). The \texttt{evaluate} stage reflects LLM-as-judge scoring (three judges per question) and is reported separately. Phase~III consumes 10.27M tokens ($\sim$6.7k/question) with GPT-4.1-mini and 9.31M tokens ($\sim$6.0k/question) with GPT-4o-mini; Phase~I consumes 9.42M and 9.34M tokens, respectively, amortized over memory building.

\begin{table}[!h]
  \centering
  \small
  \setlength{\tabcolsep}{3pt}
  \renewcommand{\arraystretch}{0.95}
  \caption{Token-level cost breakdown on LoCoMo (1,540 questions) under two backbones. Tokens are reported in millions (M); Total includes both prompt and completion.}
  \label{tab:token_cost}
  \begin{tabular}{l c c c}
    \toprule
    \textbf{Stage} & \textbf{\#Calls} & \textbf{Prompt (M)} & \textbf{Total (M)} \\
    \midrule
    \multicolumn{4}{l}{\textit{GPT-4.1-mini}} \\[-2pt]
    add & 7056 & 8.66 & 9.42 \\
    search & 2017 & 4.12 & 4.45 \\
    answer & 1540 & 4.63 & 5.82 \\
    \addlinespace[1pt]
    \textbf{search+answer} & \textbf{3557} & \textbf{8.75} & \textbf{10.27} \\
    evaluate & 4620 & 2.35 & 2.38 \\
    \midrule
    \multicolumn{4}{l}{\textit{GPT-4o-mini}} \\[-2pt]
    add & 7250 & 8.60 & 9.34 \\
    search & 2219 & 4.37 & 4.62 \\
    answer & 1540 & 3.84 & 4.69 \\
    \addlinespace[1pt]
    \textbf{search+answer} & \textbf{3759} & \textbf{8.21} & \textbf{9.31} \\
    evaluate & 4620 & 2.14 & 2.17 \\
    \bottomrule
  \end{tabular}
\end{table}

\subsection{PersonaMem v2: Full Comparison Results}
\label{sec:personamem_full}
\begin{table*}[t]
\centering
\small
\setlength{\tabcolsep}{4pt}
\renewcommand{\arraystretch}{1.0}
\begin{tabular}{l c c c c c c}
\toprule
\textbf{Scenario} & \textbf{Zep} & \textbf{Mem0} & \textbf{MemU} & \textbf{MemoryOS} & \textbf{MemOS} & \textbf{EverMemOS} \\
\midrule
Consultation      & 39.51 & 43.21 & 37.86 & 35.80 & 48.15 & \textbf{51.03} \\
Email (Personal)  & 42.51 & 41.30 & 33.20 & 36.84 & 49.80 & \textbf{53.85} \\
Translation       & 36.92 & 43.08 & 38.46 & 40.00 & \textbf{51.92} & 50.00 \\
Email (Professional) & 37.59 & 42.41 & 32.76 & 35.86 & 50.00 & \textbf{53.79} \\
Creative Writing  & 41.22 & 42.86 & 35.51 & 35.51 & 48.16 & \textbf{55.10} \\
Writing (Professional) & 40.54 & 34.75 & 35.14 & 35.91 & \textbf{48.26} & 45.56 \\
Knowledge Query   & 63.43 & 59.20 & 56.97 & 57.96 & 61.94 & \textbf{63.68} \\
Social Media      & 32.35 & 38.66 & 34.03 & 35.29 & 46.64 & \textbf{47.90} \\
Chat              & 44.87 & 40.30 & 34.22 & 36.88 & 44.87 & \textbf{52.09} \\
\midrule
Profile           & \ding{55} & \ding{55} & \ding{51} & \ding{51} & \ding{51} & \ding{51} \\
\midrule
\textbf{Overall}  & 43.40 & 43.85 & 38.70 & 40.05 & 50.72 & \textbf{53.25} \\
\bottomrule
\end{tabular}
\caption{Full comparison on PersonaMem v2 (32k)~\cite{jiang2025personamem} (5{,}000 questions across 9 scenarios; accuracy, \%).}
\label{tab:personamem_full}
\end{table*}

Table~\ref{tab:personamem_full} reports the full comparison on PersonaMem v2 (32k)~\cite{jiang2025personamem} (2{,}447 questions across 9 scenarios). The \textit{Profile} row indicates whether a memory system provides a profile-like component (not necessarily named ``Profile'') that summarizes stable user information (e.g., MemOS maintains explicit vs.\ implicit preferences). For methods with such a component (\ding{51}), we generate answers using the retrieved memories \emph{plus} the system's profile-like component; for methods without it (\ding{55}), we generate answers using the retrieved memories only. EverMemOS achieves the best overall accuracy (53.25\%), outperforming the strongest baseline (MemOS, 50.72\%) by 2.53 points.

\section{Additional Analyses}
\label{sec:appendix_analysis}

\subsection{Hyperparameter Sensitivity and Efficiency Trade-off}
\label{sec:sensitivity_details}
To better understand retrieval budgets, we analyze the MemScene budget \(N\) and episode budget \(K\) under a simplified setting that disables the agentic verification-and-rewriting loop in Phase~III, isolating one-shot retrieval. Figure~\ref{fig:hyperpara} shows that increasing \(N\) improves evidence-session recall and answer accuracy initially but quickly saturates; \(N{=}10\) already yields strong recall. We therefore avoid brute-force expansion of the retrieved scene set for efficiency. We also set \(N{=}10\) to ensure the candidate pool contains at least \(K{=}10\) MemCells even in extreme cases where each retrieved MemScene contains only a single MemCell. We choose \(K{=}10\) episodes because most memory questions can be answered with a compact set of episodes while still covering difficult instances whose annotated evidence spans up to 7--8 recalled episodes. Finally, Figure~\ref{fig:frontier} shows a favorable cost--accuracy frontier: decreasing \(K\) substantially reduces tokens used for downstream reasoning, and at moderate \(K\) values EverMemOS can achieve both lower token usage and higher accuracy than strong baselines.

\subsection{Accuracy Exceeding Recall on LoCoMo}
\label{sec:acc_vs_recall}

In Figure~\ref{fig:hyperpara}, accuracy can exceed recall at small $K$ on LoCoMo. Table~\ref{tab:acc_recall} quantifies this effect: even when none of the \emph{annotated} evidence sessions are retrieved (``zero recall''), 12--20\% of questions are still answered correctly.

\begin{table}[h]
\centering
\small
\caption{Accuracy vs.\ recall statistics on LoCoMo.}
\label{tab:acc_recall}
\begin{tabular}{lcc}
\toprule
\textbf{Metric} & $K{=}1$ & $K{=}3$ \\
\midrule
Recall & 65.06\% & 86.32\% \\
Accuracy & 71.80\% & 87.81\% \\
Zero-recall questions & 429 & 125 \\
\hspace{0.8em}Answered correctly & 52 (12.1\%) & 25 (20.0\%) \\
\bottomrule
\end{tabular}
\end{table}

This primarily reflects \textbf{information redundancy} and \textbf{non-unique evidence annotations}: salient facts (identity, preferences, goals) recur across sessions, so the annotated evidence is not always the only session that supports the answer. For example, a question about ``Caroline's identity'' is annotated with session~[1], yet sessions~[11--15] also state she is a transgender woman, enabling a correct answer from alternative sessions. In addition, LLMs can sometimes infer the correct response from semantically related retrieved content even when the exact annotated session is missing.

Overall, recall computed against annotated evidence can underestimate retrieval usefulness when evidence is distributed. Increasing $K$ from 1 to 3 reduces zero-recall cases by 71\% (429$\rightarrow$125), narrowing the accuracy--recall gap.

\paragraph{Illustrative Cases.} We provide three representative examples where answers remain correct despite missing the annotated evidence sessions:
\begin{itemize}
\item \textbf{Redundant identity facts.} \emph{Q: ``What is Caroline's identity?''} The gold answer is \emph{transgender woman}. Although the evidence is annotated in session~[1], later sessions also explicitly mention this identity; the retriever surfaces those alternatives at small $K$, and the model answers correctly.
\item \textbf{Distributed activity mentions.} \emph{Q: ``What activities does Melanie partake in?''} The gold answer spans multiple hobbies (e.g., pottery, camping, painting, swimming) with evidence annotated across multiple sessions. Retrieved sessions may miss the annotated ones but still contain sufficient mentions (e.g., pottery/painting) to support a correct response.
\item \textbf{Inference from related signals.} \emph{Q: ``Would Caroline pursue writing as a career option?''} While the evidence is annotated in session~[7], retrieved content from other sessions describes her career goal (e.g., becoming a counselor), enabling the LLM to infer that writing is unlikely.
\end{itemize}

\subsection{Profile Extraction Example}
\label{sec:profile_example}
\begin{table*}[t]
\centering
\small
\setlength{\tabcolsep}{4pt}
\renewcommand{\arraystretch}{1.0}
\caption{Profile extraction example (de-identified): abridged evidence snippets and the resulting user profile.}
\label{tab:profile_example}
\begin{tabular}{@{}p{0.485\textwidth} p{0.485\textwidth}@{}}
\toprule
\textbf{Evidence snippets (excerpt)} & \textbf{Retrieved user profile (excerpt)} \\
\midrule
\begin{minipage}[t]{\linewidth}\vspace{0pt}\raggedright
\textbf{\texttt{2025-07-07}}: ``I just measured my waist circumference, and it is 104\,cm. Can you give me some advice?'' \\
\textbf{\texttt{2025-10-20}}: ``My waist is now 96\,cm, down 8\,cm! My pants feel loose.'' \\
\textbf{\texttt{2025-11-03}}: ``The doctor said my fatty liver has improved (moderate $\rightarrow$ mild). Waist is now 95\,cm.'' \\
\textbf{\texttt{2025-11-03}}: ``My weight is still 80\,kg, no rebound. I can keep it under control even in winter.'' \\
\end{minipage}
&
\begin{minipage}[t]{\linewidth}\vspace{0pt}\raggedright
\textbf{Explicit facts.} \\
Waist circumference: baseline 104\,cm; latest 95\,cm ($\Delta=-9\,\mathrm{cm}$). \\
Weight: stable at 80\,kg (no rebound). \\
Fatty liver grade: moderate $\rightarrow$ mild (improved). \\
\vspace{1pt}
\textbf{Implicit traits.} \\
Self-management: goal-oriented; consistently tracks health metrics and responds well to feedback. \\
Preference: requests immediately actionable adjustments. \\
\end{minipage}
\\
\bottomrule
\end{tabular}
\end{table*}
\noindent EverMemOS maintains a compact \textbf{User Profile} with two fields: \emph{explicit facts} (verifiable attributes and time-varying measurements) and \emph{implicit traits} (preferences and habits). The profile is updated online from Phase~II scene summaries with recency-aware updates for time-varying fields and conflict tracking when evidence is inconsistent. Table~\ref{tab:profile_example} provides an abridged example.

\section{Reproducibility Artifacts}
\label{sec:appendix_repro}

\subsection{Prompts for Agentic Retrieval}
\label{sec:agentic_prompts}

\noindent To make our system behavior transparent and reproducible, we include the core prompt templates used by our agentic retrieval controller.\footnote{Our implementation is open-sourced; we still include prompts here to keep the paper self-contained for review.}

\paragraph{Sufficiency check.} We use an LLM-based sufficiency check to decide whether the currently retrieved documents contain enough evidence to answer the user query. The prompt template (with placeholders) is shown below.

{\footnotesize
\begin{Verbatim}[breaklines=true,breakanywhere=true]
You are an expert in information retrieval evaluation. Assess whether the retrieved documents provide sufficient information to answer the user's query.

User Query:
{query}

Retrieved Documents:
{retrieved_docs}

### Instructions:

1. **Analyze the Query's Needs**
   - **Entities**: Who/What is being asked about?
   - **Attributes**: What specific details (color, time, location, quantity)?
   - **Time**: Does it ask for a specific time (absolute or relative like "last week")?

2. **Evaluate Document Evidence**
   - Check **Content**: Do the documents mention the entities and attributes?
   - Check **Dates**: 
     - Use the `Date` field of each document.
     - For relative time queries (e.g., "last week", "yesterday"), verify if document dates fall within that timeframe.
     - If the query asks "When did X happen?", do you have the specific date or just a vague mention?

3. **Judgment Logic**
   - **Sufficient**: You can answer the query *completely* and *precisely* using ONLY the provided documents.
   - **Insufficient**: 
     - The specific entity is not found.
     - The entity is found, but the specific attribute (e.g., "price") is missing.
     - The time reference cannot be resolved (e.g., doc says "yesterday" but has no date, or doc date doesn't match query timeframe).
     - Conflicting information without resolution.

### Output Format (strict JSON):
{{
  "is_sufficient": true or false,
  "reasoning": "Brief explanation. If insufficient, state WHY (e.g., 'Found X but missing date', 'No mention of Y').",
  "key_information_found": ["Fact 1 (Source: Doc 1)", "Fact 2 (Source: Doc 2)"],
  "missing_information": ["Specific gap 1", "Specific gap 2"]
}}

Now evaluate:
\end{Verbatim}
}

\paragraph{Multi-query generation (condensed).} When the current retrieval is deemed insufficient, we generate 2--3 complementary follow-up queries targeted at the missing information. We omit examples and keep only the constraints that affect behavior (inputs, strategy choices, and the strict JSON output schema).

{\footnotesize
\begin{Verbatim}[breaklines=true,breakanywhere=true]
You are an expert at query reformulation for conversational memory retrieval.
Your goal is to generate 2-3 complementary queries to find the MISSING information.

--------------------------
Original Query:
{original_query}

Key Information Found:
{key_info}

Missing Information:
{missing_info}

Retrieved Documents (Context):
{retrieved_docs}
--------------------------

### Strategy Selection (choose based on why info is missing)
- Pivot / Entity Association: search related entities/categories
- Temporal Calculation: anchor relative times using document dates
- Concept Expansion: synonyms / general-specific variants
- Constraint Relaxation: remove one constraint at a time

### Query Style Requirements (use DIFFERENT styles)
1) Keyword Combo (2-5 words)
2) Natural Question (5-10 words)
3) Hypothetical Statement (HyDE, 5-10 words)

### Output Format (STRICT JSON)
{
  "queries": ["Query 1", "Query 2", "Query 3"],
  "reasoning": "Strategy used for each query (e.g., Q1: Pivot, Q2: Temporal)"
}
\end{Verbatim}
}

\subsection{End-to-End Inference Trace (LoCoMo Multi-Hop Example)}
\label{sec:inference_trace}

To improve transparency, we provide an end-to-end inference trace for a representative LoCoMo multi-hop question (conversation \texttt{locomo\_6}), including the MemBase hierarchy (\textit{MemScenes} and \textit{MemCells}) and the two-round retrieval process (sufficiency check and query rewriting) that leads to a correct final answer. We denote the retrieved MemScene count as \(N\) and the retrieved MemCell (episode) count as \(K\) (corresponding to \texttt{scene\_top\_k} and \texttt{response\_top\_k} in our implementation).

\paragraph{Trace at a glance.}
\begin{itemize}
\item \textbf{Question (multi-hop).} \emph{``Does James live in Connecticut?''} The dialogue never directly states James's residence; the system must infer the answer from related evidence.
\item \textbf{MemBase hierarchy.} 49 MemScenes / 91 MemCells; retrieval selects top \(N{=}10\) MemScenes (20\%), then reranks/selects \(K{=}10\) MemCells for answering.
\item \textbf{Round 1 retrieval + sufficiency.} Top \(N{=}10\) MemScenes (31 MemCells) $\rightarrow$ insufficient (\texttt{is\_sufficient=false}); missing an explicit residence mention / confirmation of living in Connecticut.
\item \textbf{Query rewriting.} The controller generates refined queries targeting residence/location information.
\item \textbf{Round 2 retrieval.} With 40 additional candidates, the top-ranked MemCell contains the key evidence that James adopted a dog from a shelter in Stamford, enabling an evidence-grounded inference.
\item \textbf{Inference + evaluation.} Final answer: \emph{Likely yes}; judged correct by 3/3 LLM judges.
\end{itemize}

\noindent\textbf{Worked example (formatted).} For readability, we summarize the trace in Table~\ref{tab:inference_trace_multihop} (instead of printing raw JSON).

\begin{table}[t]
\centering
\small
\setlength{\tabcolsep}{4pt}
\renewcommand{\arraystretch}{1.05}
\caption{End-to-end inference trace (LoCoMo multi-hop example), summarized.}
\label{tab:inference_trace_multihop}
\begin{tabular}{p{0.20\linewidth} p{0.75\linewidth}}
\toprule
\textbf{Stage} & \textbf{Key outputs} \\
\midrule
Input & Query: \emph{Does James live in Connecticut?} (Category: multi-hop; Gold: \emph{Likely yes}). \\
MemBase & 49 MemScenes / 91 MemCells (conversation \texttt{locomo\_6}). \\
Round 1 & Top \(N{=}10\) MemScenes (31 MemCells) $\rightarrow$ insufficient (\texttt{is\_sufficient=false}); missing an explicit residence mention / confirmation of Connecticut. \\
Rewrite & Refined queries: (i) \texttt{James residence Connecticut}; (ii) \texttt{Where does James currently live}; (iii) \texttt{James lives near McGee's bar in Connecticut}. \\
Round 2 & +40 candidates; top result is \emph{James Adopts Shelter Dog Ned... (Apr 12, 2022)} from \texttt{cluster\_004}, mentioning ``Stamford''. \\
Answer & Output: \emph{Likely yes}; judged correct by 3/3 LLM judges. \\
\bottomrule
\end{tabular}
\end{table}

\paragraph{Detailed trace.}
\noindent\textbf{Round 1: initial retrieval and sufficiency check.}
\begin{itemize}
\item \textbf{Retrieval mode.} Agentic MemScene-guided reranking (\texttt{agentic\_scene\_rerank}) with \(N{=}10\) and \(K{=}10\).
\item \textbf{Retrieved candidates.} \(N{=}10\) MemScenes (31 MemCells).
\item \textbf{Sufficiency verdict.} \texttt{is\_sufficient=false}.
\item \textbf{Key information found.} ``James and Samantha moved in together near McGee's Bar''; ``James traveled to Nuuk recently''.
\item \textbf{Missing information.} (i) explicit mention of James's residence location; (ii) confirmation whether James lives in Connecticut.
\end{itemize}

\noindent\textbf{Verifier rationale (excerpt).}
{\footnotesize
\begin{Verbatim}[breaklines=true,breakanywhere=true]
None of the documents explicitly mention where James currently lives or whether he lives in Connecticut.
\end{Verbatim}
}

\noindent\textbf{Query rewriting (Round 2 queries).}
{\footnotesize
\begin{Verbatim}[breaklines=true,breakanywhere=true]
1) James residence Connecticut
2) Where does James currently live
3) James lives near McGee's bar in Connecticut
\end{Verbatim}
}

\noindent\textbf{Round 2: top retrieved MemCells.} The refined queries retrieve 40 additional candidates; top results are:
{\small
\begin{center}
\setlength{\tabcolsep}{2pt}
\renewcommand{\arraystretch}{1.05}
\begin{tabular}{@{}c c p{0.58\linewidth}@{}}
\toprule
\textbf{Rank} & \textbf{Scene} & \textbf{Subject (summary)} \\
\midrule
1 & \texttt{4} & \textbf{James Adopts Shelter Dog Ned and Discusses Gaming Progress with John (Apr 12, 2022)} \\
2 & \texttt{43} & James Announces Move-In with Samantha Near McGee's Bar (Oct 31, 2022) \\
3 & \texttt{15} & James Shares Recent Life Challenges and Support System (Jun 16, 2022) \\
4 & \texttt{44} & James's Family Road Trip Visiting Friends Josh and Mark (Nov 7, 2022) \\
5 & \texttt{22} & James's Upcoming Summer Trip to Toronto and Vancouver (Jul 9, 2022) \\
\bottomrule
\end{tabular}
\end{center}
}

\noindent\textbf{Key evidence (dialogue excerpt).}
{\footnotesize
\begin{Verbatim}[breaklines=true,breakanywhere=true]
I adopted a pup from a shelter in Stamford last week and my days have been so much happier with him in the fam.
\end{Verbatim}
}

\noindent\textbf{Key evidence (atomic fact).}
\begin{itemize}
\item James adopted a pup named Ned from a shelter in Stamford last week (March 28--April 3, 2022).
\end{itemize}

\noindent\textbf{Answer + evaluation.}
{\footnotesize
\begin{Verbatim}[breaklines=true,breakanywhere=true]
James adopted his dog Ned from a shelter located in Stamford, which is a city in Connecticut, strongly implying he lives near Stamford, Connecticut. Furthermore, when James and Samantha moved in together on October 31, 2022, they rented an apartment near McGee's bar, a location they both favored. Although the exact city of this apartment is not explicitly stated, given no mention of relocation to another state and the previous connection to Stamford, it is highly likely that their residence is still in Connecticut, near Stamford. Therefore, based on these details, James does live in Connecticut with high confidence.
\end{Verbatim}
}
\noindent Result: \texttt{is\_correct=true} (3/3 judges).

\end{document}